\documentclass[preprint]{article}

\usepackage[nonatbib]{neurips_2022} % regular references, [nonatbib]

\usepackage[utf8]{inputenc} % allow utf-8 input
\usepackage[T1]{fontenc}    % use 8-bit T1 fonts
\usepackage{hyperref}       % hyperlinks
\usepackage{url}            % simple URL typesetting
\usepackage{booktabs}       % professional-quality tables
\usepackage{amsfonts}       % blackboard math symbols
\usepackage{nicefrac}       % compact symbols for 1/2, etc.
\usepackage{microtype}      % microtypography
\usepackage{xcolor}         % colors

\usepackage{listings}       % code
\usepackage{graphicx}       % pictures
\usepackage{multirow}

\definecolor{codegreen}{rgb}{0,0.6,0}
\definecolor{codegray}{rgb}{0.5,0.5,0.5}
\definecolor{codepurple}{rgb}{0.58,0,0.82}
\definecolor{backcolour}{rgb}{0.95,0.95,0.92}

\lstdefinestyle{mystyle}{
    backgroundcolor=\color{backcolour},   
    commentstyle=\color{codegreen},
    keywordstyle=\color{magenta},
    numberstyle=\tiny\color{codegray},
    stringstyle=\color{codepurple},
    basicstyle=\ttfamily\footnotesize,
    breakatwhitespace=false,         
    breaklines=true,                 
    captionpos=b,                    
    keepspaces=true,                 
    numbers=left,                    
    numbersep=5pt,                  
    showspaces=false,                
    showstringspaces=false,
    showtabs=false,                  
    tabsize=2
}

\title{ESTA: An Esports Trajectory and Action Dataset}

\author{%
  Peter Xenopoulos \\
  New York University\\
  \texttt{xenopoulos@nyu.edu} \\
  \And
  Claudio Silva \\
  New York University\\
  \texttt{csilva@nyu.edu} \\
}

\begin{document}

\maketitle

\begin{abstract}
Sports, due to their global reach and impact-rich prediction tasks, are an exciting domain to deploy machine learning models. However, data from conventional sports is often unsuitable for research use due to its size, veracity, and accessibility. To address these issues, we turn to esports, a growing domain that encompasses video games played in a capacity similar to conventional sports. Since esports data is acquired through server logs rather than peripheral sensors, esports provides a unique opportunity to obtain a massive collection of clean and detailed spatiotemporal data, similar to those collected in conventional sports. To parse esports data, we develop \texttt{awpy}, an open-source esports game log parsing library that can extract player trajectories and actions from game logs. Using \texttt{awpy}, we parse 8.6m actions, 7.9m game frames, and 417k trajectories from 1,558 game logs from professional Counter-Strike tournaments to create the Esports Trajectory and Actions (ESTA) dataset. ESTA is one of the largest and most granular publicly available sports data sets to date. We use ESTA to develop benchmarks for win prediction using player-specific information. The ESTA data is available at \href{https://github.com/pnxenopoulos/esta}{https://github.com/pnxenopoulos/esta} and \texttt{awpy} is made public through PyPI.

% graphs, trajectory prediction, win probability
% rl applications?
% game with asymmetric information
% no trajectory or action data set as big. collection of movements and actions in a well-defined space
%  The abstract paragraph should be indented \nicefrac{1}{2}~inch (3~picas) on
%  both the left- and right-hand margins. Use 10~point type, with a vertical
%  spacing (leading) of 11~points.  The word \textbf{Abstract} must be centered,
%  bold, and in point size 12. Two line spaces precede the abstract. The abstract
%  must be limited to one paragraph.
% Sports, due to their wide appeal, well-defined rules, and slew of impact-rich applications, are an exciting domain for machine learning research.
\end{abstract}

\section{Introduction} \label{sec:introduction}
Sports are an exciting domain for machine learning due to their global appeal, well-defined rules, and multitude of machine learning tasks, such as win probability estimation, trajectory prediction, and tactic identification. Especially in the last decade, sports analytics has garnered much interest from teams, leagues, and fans~\cite{DBLP:journals/bigdata/AssuncaoP19}. However, while conventional sports, like soccer, basketball, and baseball, produce large and varied data, such as videos or player locations, the data suffer from various drawbacks that hamper its usability in research~\cite{DBLP:journals/csur/GudmundssonH17}.

Most conventional sports tracking data is acquired using player-embedded sensors or computer vision-based approaches~\cite{DBLP:journals/csur/GudmundssonH17}. While the former often produces clean data, it is difficult to scale sensor-based tracking to acquire large amounts of data, due to factors such as privacy concerns from players or financial constraints in acquiring and calibrating a sufficient amount of sensors~\cite{rana2020wearable}. Computer vision-based techniques often require significant cleaning and complex computational workflows~\cite{DBLP:journals/tomccap/StenslandGTHNAMLLLGHSJ14, DBLP:conf/ism/TennoeHNASGJGH13}. Furthermore, one must collect the video themselves, which imposes further time and cost constraints.

Sports data is also often private or hard to access. Oftentimes, this is due to business factors, such as data exclusivity agreements in professional leagues which prohibit data sharing to organizations outside the league~\cite{longenhagen_mcdaniel_2019, streeter_2019}. Additionally, access to professional sports organizations and players is seldom granted to the public which prohibits researchers from sharing data~\cite{socolow2017game}. While some tracking data has been released for some sports, it is not well-maintained nor well-documented. Furthermore, the data is oftentimes small and suitable mostly for demo use, which limits its usability for machine learning research.

Esports offer a unique opportunity to produce clean and granular sports data at scale~\cite{DBLP:journals/intr/HamariS17}. In conventional sports, optics and sensors may be directed towards a playing surface, which capture player locations and in-game events. In esports, players connect to a game server to play a video game. While there is no physical world to monitor, such as through peripheral sensors or optics required by conventional sports, the game server generates a server log. Server-side logs provide the ability to reconstruct a game, and the game server typically writes to the log at a rate of dozens of times per second. Thus, esports data is not only of high quality but can also be collected at a high frequency. Furthermore, esports game logs are often publicly available, since players, leagues, and tournaments routinely upload server logs for verification and playback purposes. 

Some popular esports, like StarCraft II, Defense of the Ancients 2, and League of Legends, are played from an isometric perspective. Counter-Strike: Global Offensive (CSGO), however, is played in a first-person capacity. Thus, a player navigates through the map from the first-person view of their in-game agent, as opposed to navigation from a top-down view. In this sense, first-person esports like CSGO are quite similar to conventional sports, particularly in the data that they generate. Thus, a large spatiotemporal dataset derived from esports would not only be useful for the esports analytics community, but also for the general sports analytics community, given that prediction tasks, like win prediction, are shared across sports~\cite{DBLP:conf/bigdataconf/XenopoulosDS20, DBLP:conf/bigdataconf/XenopoulosS21, DBLP:conf/www/XenopoulosFS22, DBLP:conf/aist/MakarovSLI17, yurko2020going, DBLP:conf/kdd/RobberechtsHD21}. In this work, we make the following contributions:

\begin{enumerate}
    \item \textbf{\texttt{awpy} Python library.} We introduce \texttt{awpy}, a Python library to parse, analyze, and visualize Counter-Strike game replay files. The parsed JSON output contains game replay, server, and parser metadata, along with a list of ``game round'' objects that contain player actions and locations. Awpy is able to accept user-specified parsing arguments to control the parsed data. We make awpy available on PyPI through \texttt{pip install awpy}.
    \item \textbf{Trajectories and Actions dataset.} Using awpy, we create the \textbf{ES}ports \textbf{T}rajectories and \textbf{A}ctions (ESTA) dataset. ESTA contains parsed demo and match information from 1,558 professional Counter-Strike games. ESTA contains 8.6m player actions, 7.9m total frames, and 417k player trajectories. ESTA represents one of the largest publicly available sports datasets where tracking and event data are coupled. ESTA is made available through Github.
    \item \textbf{Benchmark tasks.} Using the ESTA data, we provide benchmarks for sports outcome prediction, and in particular, win probability prediction~\cite{DBLP:conf/kdd/RobberechtsHD21, DBLP:conf/bigdataconf/XenopoulosS21}. Win probability prediction is a fundamental prediction task in sports and has multiple applications, such as player valuation~\cite{DBLP:conf/bigdataconf/XenopoulosDS20, DBLP:conf/kdd/SiciliaPG19, DBLP:conf/kdd/DecroosBHD19}, and game understanding~\cite{DBLP:conf/www/XenopoulosFS22, DBLP:conf/kdd/PowerRWL17}.
\end{enumerate}

% ~\cite{DBLP:conf/cvpr/AlahiGRRLS16, DBLP:conf/cvpr/GuptaJFSA18, DBLP:conf/cvpr/SadeghianKSHRS19, DBLP:conf/eccv/YuMRZY20}

% data could be useful for research sequences/RL/etc., for teaching, for visualization. To this end, we make the following contributions:

% . Second, sports data is often private

% talk about how sports data has video, images, trajectories, actions,

% The first issue is that data acquisition can be flawed, due to reliance on object tracking and so on....

% Secondly, this data is often private, either exclusive to particular sports teams or leagues, or not maintained or properly documented. 

% Lastly, in cases that detailed sports data exist publicly, the data is oftentimes small and suitable mostly for demo use.

% To address these issues, we turn to esports, or competitive video gaming. Due to the computer-based nature of esports, the data is easy to acquire, accurate and granular. 

% to which one can apply machine-learning techniques due to its global appeal, well-defined rules
\section{Related Work} \label{sec:related-work}
Spatiotemporal data in sports, particularly in the form of player actions or player trajectories, has garnered much interest. A common task is to predict an outcome in some period of a given game. To do so, the prediction model typically takes a ``game state'' as input, which contains the context of the game at prediction time. For example, Decroos~et~al. represent a soccer game as a series of on-the-ball actions~\cite{DBLP:conf/kdd/DecroosBHD19}. They define a game state as the previous three actions, and predict if a goal will occur in the next ten actions. For esports, Xenopoulos~et~al. perform a similar procedure, but rather define a game state as a snapshot of global and team-specific information at prediction time~\cite{DBLP:conf/bigdataconf/XenopoulosDS20}. In each of the aforementioned works, event or trajectory data was processed to create spatial features used in prediction, and the game state was represented by a vector.

Game states are increasingly being defined in structures which require techniques such as recurrent or graph neural networks. For example, Yurko~et~al. use a sequence of game states as input to a recurrent model, to predict the total yards gained in an American football play~\cite{yurko2020going}. Sicilia~et~al. train a multiclass sequence prediction model to predict the outcomes of a basketball possession~\cite{DBLP:conf/kdd/SiciliaPG19}. Their unit of concern is a basketball possession, which they define as a sequence of $n$ moments, each described by player locations. Xenopoulos~et~al. define a game's context using a graph representation, where players are nodes in a fully connected graph~\cite{DBLP:conf/bigdataconf/XenopoulosS21}.

Predicting player movement, rather than a specific sports outcome, is also a common task in sports. Yeh~et~al. proposed a graph variational recurrent neural network approach to predict player movement in basketball and soccer~\cite{DBLP:conf/cvpr/YehSH019}. Falsen~et~al. use a conditional variance autoencoder to create a generative model which predicts basketball player movement~\cite{DBLP:conf/eccv/FelsenLG18}. Omidshafiei~et~al. use graph networks and variational autoencoders to impute missing trajectories~\cite{deepmind_2022}. To do so, they use a proprietary tracking dataset from the English Premier League.

Although sports data can oftentimes be large, it is rarely publicly available or well-documented, especially since many sports data acquisition systems are proprietary. A popular large and public sports dataset is a collection of around 600 games from the 2015-2016 NBA season~\cite{nba-data}. Each game contains tracking data for both the players and the ball, collected at 20Hz. The NBA dataset is often downsampled in practice. For example, Yeh~et~al., using the NBA dataset, create trajectories of 50 frames parsed at 6 Hz, which corresponds to roughly 121k trajectories in total~\cite{DBLP:conf/cvpr/YehSH019}. One downside to the NBA dataset is that it lacks a dedicated maintainer and reflects older tracking technology. Pettersen~et~al. introduce a collection of both tracking and video data for about 200 minutes of professional soccer~\cite{DBLP:conf/mmsys/PettersenJJBGMLGSH14}. The data is well-documented and contains roughly 2.5 million samples, not only including player locations but also kinetic information such as speed and acceleration. However, the data is geared towards computer vision-based tasks, rather than outcome prediction, and lacks player actions, such as passes, shots, or tackles.

There also exist a limited number of esports-specific datasets, given that some game data is easily accessible at scale. Lin~et~al. introduce \textit{STARDATA}, a dataset containing millions of game states and player actions from StarCraft: Brood War~\cite{DBLP:conf/aiide/LinGKS17}. Smerdov~et~al. propose a dataset containing not only in-game data but also physiological measurements from professional and amateur League of Legends players for around two dozen matches~\cite{smerdov-esports}. For Counter-Strike: Global Offensive (CSGO), there exists the PureSkill.gg dataset of parsed amateur matches~\cite{pureskill}. While the aforementioned dataset is large, it is hosted on AWS Data Exchange, and thus requires financial resources if the data is to be used outside of AWS resources. Furthermore, the parser used to generate the datasets is not public. Finally, one may also find esports datasets on Kaggle, however, these datasets are oftentimes undocumented, deprecated, or old, which hampers their use for reproducible research. Some video games, like Defense of the Ancients 2 (Dota 2), have a range of fan sites which also host data~\cite{opendota}. 
%For example, OpenDota, a popular Dota 2 fan site, hosts APIs and has created data dumps containing millions of parsed Dota 2 matches. 

Many esports are played from an isometric perspective. In that sense, the trajectory and action data attained in them is somewhat different than the data found in conventional sports. First-person shooter (FPS) is a popular video game genre in which a player controls their character in a first-person capacity, as opposed to the top-down view found in games like StarCraft 2 or Dota 2. Thus, the data generated by FPS-style games is more similar to conventional sports than data from StarCraft 2 or Dota 2. In this work, we focus on CSGO, a popular FPS esport. At the time of writing, CSGO achieves roughly one million daily peak players compared to Dota 2, which attains roughly 700,000 daily peak players~\cite{steamcharts}. In particular, our proposed dataset aims to address issues regarding accessibility, documentation, and size in existing esports datasets.

% Like StarCraft 2 and Dota 2, CSGO data is also available on Kaggle, however, similarly, it is often small, undocumented and lacks full game replays. 

%trajectories~\cite{DBLP:conf/icpr/MontiBCC20}
% trajectory and video~\cite{DBLP:conf/cvpr/SanfordGHPJ20}
% trajectory prediction \cite{DBLP:conf/iclr/ZhanZYSL19}
% trajectory shot tennis \cite{DBLP:journals/tkde/WeiLMS16}

% NFL Big data bowl (14193 plays, 24728130 tracking points)
% One NBA game (211445 frames with 11 points each)
% Norway 2493978 points
% https://github.com/sealneaward/nba-movement-data
% https://github.com/linouk23/NBA-Player-Movements
\section{The \texttt{awpy} package} \label{sec:awpy}
One of the main objectives of this work is to expand the public tools by which people can parse widely available esports data into a format conducive for analysis. To that end, we introduce the awpy Python library, which we use to create the ESTA dataset. In this section, we detail CSGO, its data, and the awpy library. Our library can be installed via \texttt{pip install awpy} and is available at \url{https://github.com/pnxenopoulos/awpy}. We provide example Jupyter Notebooks and detailed documentation both for awpy's functionality, as well as for its JSON output, in the linked Github repository.

\subsection{Counter-Strike Background} \label{sec:csgo-background}
Counter-Strike is a long-running video game series, with the most recent rendition, CSGO, attracting roughly one million peak users at time of writing. CSGO has a robust competitive scene, with organized tournaments occurring year-round both in-person (denoted LAN for local area network) and online. CSGO is a round-based game, whereby two teams of five players each attempt to win each round by reaching a win condition. Teams play as one of two ``sides'', denoted CT and T, and then switch sides after fifteen rounds. The T side can win a round by eliminating members of the opposing side, or by planting a bomb at one of two bombsites. The CT side can win a round by eliminating members of the opposing side, defusing the bomb, or by preventing the T side from planting the bomb in the allotted round time. Rounds last around one to two minutes each.

Broadly, there are four different phases of play. When a new round starts, a game phase called ``freeze time'' begins. In this phase, which lasts 20 seconds, players are frozen in place and may not move, but they may buy equipment, such as guns, grenades, and armor using virtual money earned from doing well in previous rounds. The next phase of the game is the ``default'' game phase where players can move around and work towards one of their side's win conditions. This game phase, which may last up to one minute and 55 seconds, constitutes the majority of a round. The default game phase ends when either team is eliminated, the clock time runs out, or the bomb is planted. When the bomb is planted, the ``bomb'' phase begins and lasts 40 seconds. During this time, the T side (which planted the bomb) can still win the game by eliminating all CT players. Once a win condition has been met, regardless of the previous phase (default or bomb planted), the ``round end'' phase begins, in which players can still move around and interact for five seconds until a new round begins. 

Players start each round with 100 health points (HP) and are eliminated from a round when they reach 0 HP. Players lose HP when they are damaged -- typically from gunfire and grenades (also called utility) from the opposing side. Each CSGO game takes place in a virtual world called a ``map''. Competitive CSGO events typically have seven standard maps on which players may play. At the beginning of a competitive match, the two competing teams undergo a process to determine which map(s) they will play. Typically, competitive CSGO matches are structured as a best-of-one, best-of-three, or rarely, best-of-five maps.

\subsection{Counter-Strike Data} \label{sec:csgo-data}
In multiplayer video games, players (clients) connect to a game server. When players use peripherals, like a mouse or keyboard, they change their local ``game state'', which is what the player sees on their screen. The clients send these changes to the server, which reconciles client inputs from connected players and returns a unified, global game state to clients. These updates between the clients and game server happen at a predefined \textit{tick rate}, which is usually 128 in competitive play, meaning the clients and server update 128 times a second. In fact, the game server dispatches ``events'', which are game state deltas that the client uses to construct the game state.

\begin{figure}
    \centering
    \includegraphics[width=0.8\textwidth]{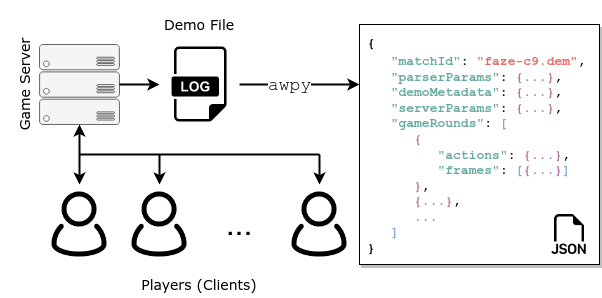}
    \caption{Players (clients) send updates to the game server, which returns a unified global game state back to clients through game event dispatches. This process occurs over 100 times a second. As the game server sends updates to clients, it records these event dispatches to a log called a demo file. The awpy Python library allows a user to parse CSGO demo files into a JSON structure which contains player actions and locations.}
    \label{fig:parsing-csgo-file}
\end{figure}

The game server records the aforementioned updates and saves a file called a demo file (colloquially known as a demo), which is effectively a serialization of the client-server network communication. Demo files allow one to recreate the game as it happened from the game server's point-of-view, and can be loaded in the CSGO application itself to view past matches. A demo file is restricted to the performance on a single map. Thus, a competitive CSGO match may produce multiple demos. Bednarek~et~al. describe CSGO demo files in further detail~\cite{DBLP:conf/data/BednarekKYZ17}. CSGO demo files are easy to acquire, and can be found in the game itself, or on many third-party matchmaking, tournament, and fan sites. Thus, CSGO demo files constitute an abundant source of multi-agent interaction data. We show an example of demo file generation and parsing made possible by awpy in Figure~\ref{fig:parsing-csgo-file}.

\subsection{Using awpy to parse CSGO demos}
While there exist a few CSGO demo file parsers, they are often written in languages that are uncommon to many data science or machine learning workflows, such as Go~\cite{parser-golang}, Node.js~\cite{parser-node}, or C\#~\cite{parser-csharp}. Furthermore, one must often extend upon the aforementioned libraries in order to output the data into a commonly used format, such as CSV or JSON. Thus, we created awpy to provide simple CSGO parsing functionality in Python. We show an example of awpy in practice in Listing~\ref{listing:awpy_example}. There are four main pieces of information in the parsed JSON output. First, many top-level keys indicate match and demo metadata, such as the map, tick rate, or total number of frames. Second, the ``parserParameters'' key contains the parameters used to parse the demo file, such as the user-specified rate at which game snapshots are parsed. Third, ``serverVars'' contains server-side information on game rules, such as the round length or bomb timer. Lastly, ``gameRounds'' is a list containing information on each round of the game. While each element in ``gameRounds'' contains round metadata, like the starting equipment values or the round end reason, it also contains lists of player events and lists of game frames, which we discuss in detail in Section~\ref{sec:dataset}.

In addition to parsing, awpy also contains functions to calculate player statistics or visualize player trajectories and actions. The awpy library hosts four modules: (1) analytics, (2) data, (3) parser, and (4) visualization. The analytics module contains functions to generate summary statistics for a single demo. The data module contains useful data on each competitive map's geometry and navigation meshes, which are used by in-game computer-controlled bots to move in-game. The parser module contains the functions to both parse CSGO demo files and clean the output. Although CSGO demos may contain errant rounds or incorrect scores, usually caused by third-party server plugins, they are relatively straightforward to identify and remedy by following CSGO game logic, which awpy's parse module addresses. Finally, the visualization module contains functions to plot frame and round data in both static and dynamic fashion.

\begin{lstlisting}[language=Python,caption={awpy can be used to parse, analyze, and visualize CSGO demo files.},captionpos=b]
# pip install awpy
from awpy.parser import DemoParser
from awpy.analytics.stats import player_stats
from awpy.visualization.plot import plot_round

# Parse the demo
p = DemoParser(demofile="faze-vs-cloud9.dem", parse_rate=128)
demo_data = p.parse()

# Analyze and aggregate player statistics
player_data = player_stats(demo_data["gameRounds"])

# Visualize the first frame of the first round
frame = demo_data["gameRounds"][1]["frames"][0]
map_name = demo_data["mapName"]
plot_frame(frame, map_name)
\end{lstlisting}
\label{listing:awpy_example}
\section{The ESTA Dataset} \label{sec:dataset}
The \textbf{ES}ports \textbf{T}rajectories and \textbf{A}ctions (ESTA) dataset contains parsed CSGO game demos. Each parsed demo is a compressed JSON file and contains granular player actions (Section~\ref{sec:actions}) and game frames (Section~\ref{sec:frames}). Game frames, which are effectively game ``snapshots'', contain all game information at a given time, and are parsed at a rate of 2 frames per second. Each JSON file also contains details on six player action types: damages, kills, flashes, bomb plants, grenades, and weapon fires. ESTA is released under a CC BY-SA 4.0 license\footnote{ESTA is available at \url{https://github.com/pnxenopoulos/esta}}.

The ESTA dataset includes games from important CSGO tournaments. Each of these tournaments was held in a local area network (LAN) environment between January 2021 and May 2022. We collated the list of tournaments using \url{https://www.hltv.org/}, a popular CSGO fan site. From this list of tournaments and their associated matches, we obtained the corresponding demo files and parsed them using awpy. The players involved in each parsed demo, as well as the demo files themselves, are already public. Therefore, there are no privacy concerns with regards to player data. We disregarded parsed demos which had an incorrect number of parsed rounds. In total, ESTA contains 1,558 parsed JSONs which contain 8.6m actions, 7.9m frames, and 417k trajectories. We have the following counts of demos for each map: Dust 2 (197), Mirage (278), Inferno (289), Nuke (260), Overpass (181), Vertigo (169), Ancient (132), Train (52). While Train is not part of the \textit{current} competitive map pool, it was part of the map pool for a portion of the time frame which ESTA covers, until Ancient replaced it. 

% PGL Major 2022, PGL Major 2021, ESL Pro League 15, IEM Katowice 2022, Blast Premier World Final 2021, IEM Winter 2021, Blast Premier Fall Final 2021, IEM Fall 2021 Europe, IEM Cologne 2021/Play in
% note here also total GB

% ONLINE, 2.2GB compressed, 896 demos
% Total rounds: 23444
% Total actions: 4867532
% Total frames: 4360444

% LAN, 1.7GB compressed, 692 demos
% Total rounds: 18338
% Total actions: 3,924,550
% Total frames: 3,484,090

\subsection{Player actions} \label{sec:actions}
\begin{figure}
    \centering
    \includegraphics[width=\textwidth]{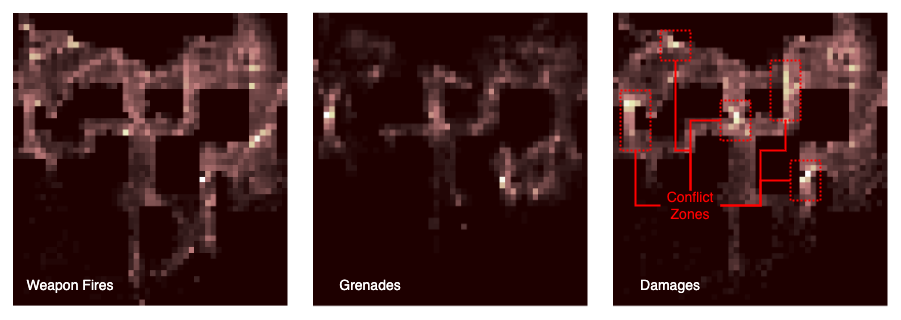}
    \caption{Heatmaps showing the shooter location for weapon fire events (left), the landing position for grenades (center), and the victim coordinates for damage events (right) on Dust 2 demo files. Bright color indicates higher action density. We can see distinct regions of high density for each action. For example, for damages, there are clear ``conflict zones'', indicating regions of the map where fights often occur.}
    \label{fig:actions-summarized}
\end{figure}

Player actions are crucial because they fundamentally change the game's state. Many of the event dispatches that the server sends are due to player actions. Broadly, player actions can be categorized as \textit{local}, if the action involves some engagement between two players, or \textit{global}, if the action has an effect on the global attributes of the game state. Examples of local actions include player damages, kills, or flashes, which involve some interaction between players. Global actions include bomb events, like plants or defuses, grenade throws, or weapon fires. These events typically change the global state by changing the bomb status (e.g. bomb plant) or by imposing temporary constraints on the map through fires or smokes that reduce visibility. On average, about 210 of the six parsed actions occur per CSGO round. Weapon fire events make up 68\% of the total actions, damages 13\%, grenade throws 10\%, flashes 6\%, kills 3\%, and bomb events less than 1\%. Each event contains locations of each player involved in the action (e.g., damages actions involve a player doing the damage, and a player receiving the damage). In Figure~\ref{fig:actions-summarized}, we show an example of actions from Dust 2 demos summarized by location. It is evident that there are areas where actions often occur (light-shaded areas), which is indicative of the common fight locations. These patterns may be leveraged by users to identify tactics or create advanced features. For example, certain grenades or fights may be indicative of a team's future intentions. 

% At the same time, due to its immense size, the ESTA data also provides many situations where players deviate. 

\subsection{Game frames} \label{sec:frames}
Game frames record the state of the game at a specific tick. Within each frame, awpy parses a combination of global, team, and player-specific information. Globally, a game frame contains temporal information, such as the tick and clock time. Additionally, each snapshot contains a list of active ``fires'' and ``smokes'', which are temporary changes to the map incurred by the players using their equipment, namely, their grenades. These may temporarily block locations or lower visibility in certain parts of the map. Lastly, information on the bomb, such as if and where it is planted, is also provided. On average, 188 game frames occur per round when parsed at a rate of 2 frames per second. Thus, the average CSGO round in ESTA lasts 94 seconds. About 30\% of rounds contain a bomb plant, and 37\% end in a CT elimination win, 12\% in bomb defusal, 5\% in target saved, 19\% in bomb exploded, and 27\% in T elimination.

\begin{figure}
    \centering
    \includegraphics[width=\textwidth]{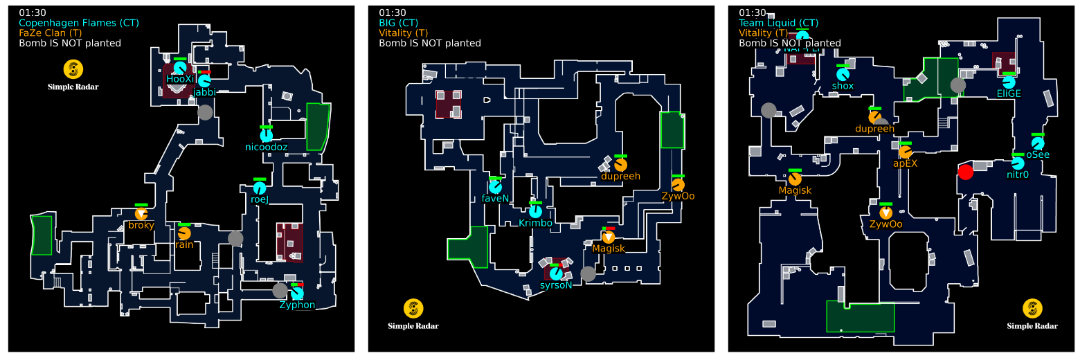}
    \caption{Game frames on demos on three maps: ``Inferno'' (left), ``Mirage'' (center), and ``Dust 2'' (right). CT players are represented by cyan and T players by orange. Above each player is a bar showing their HP. Each player also has a black line indicating the direction they are facing. The bomb is identified by a white triangle, active fires by a red circle, and active smokes by a gray circle.}
    \label{fig:game_frame}
\end{figure}

In addition to global information, game frames contain information on both the T and CT side. Namely, each team consists of a list of players, along with other team-specific attributes which are typically aggregated from the list of players, such as the number of alive players for a team. In each frame, each player is represented by over 40 different attributes, such as their location, view direction, inventory, and ping, a measure of network latency to the game server. We show a visual representation of three game frames from three different maps in Figure~\ref{fig:game_frame}. Within awpy, one can create both static (for a single frame) and animated (for a sequence of frames) visualizations.

% https://www.hltv.org/matches/2356154/copenhagen-flames-vs-faze-pgl-major-antwerp-2022
% https://www.hltv.org/matches/2356157/big-vs-vitality-pgl-major-antwerp-2022
% https://www.hltv.org/matches/2356152/vitality-vs-liquid-pgl-major-antwerp-2022

%\begin{figure}
%    \centering
%    \includegraphics[width=\linewidth]{figures/awpy_data_model.pdf}
%    \caption{ESTA contains demo and round metadata, along with player actions (blue) and frames, which contain player positions (red).}
%    \label{fig:data_model}
%\end{figure}
\section{Benchmarks} \label{sec:experiments}

\subsection{Problem Formulation}
Predicting the outcome of a game, also known as win probability prediction, is an important task for a variety of applications, such as player valuation, sports betting, fan engagement, and game understanding. As such, it is a commonly researched prediction task in the sports analytics community in both conventional sports and esports~\cite{DBLP:conf/bigdataconf/XenopoulosDS20, DBLP:conf/bigdataconf/XenopoulosS21, DBLP:conf/www/XenopoulosFS22, DBLP:conf/aist/MakarovSLI17, yurko2020going, DBLP:conf/kdd/RobberechtsHD21, DBLP:conf/kdd/DecroosBHD19}. Broadly, the goal is to predict $Y_i$, the outcome of game period $i$. To do so, past literature typically uses the state of the game time $t$, which occurs in period $i$, as input~\cite{DBLP:conf/bigdataconf/XenopoulosDS20, DBLP:conf/bigdataconf/XenopoulosS21, DBLP:conf/www/XenopoulosFS22, yurko2020going, DBLP:conf/aist/MakarovSLI17, DBLP:conf/kdd/RobberechtsHD21}. We refer to this input as $g_t$. In CSGO, we are most concerned with the outcome of a round, so we set $Y_i = 1$ when the CT side wins round $i$ and 0 otherwise. Thus, we want to train a model to predict $\mathbb{P}(Y_i = 1 \mid g_t)$, where $g_t$ is an object that describes the state of a game at time $t$. The aforementioned formulation, using a game state as input, is common when predicting other game outcomes, such as the end yardline of a play in American football or the probability of scoring in soccer~\cite{yurko2020going, DBLP:conf/kdd/DecroosBHD19}.

\subsection{Representing Game States}
There are many ways to represent $g_t$, such as through a single vector containing aggregated team information~\cite{DBLP:conf/bigdataconf/XenopoulosDS20, DBLP:conf/www/XenopoulosFS22} or graphs~\cite{DBLP:conf/bigdataconf/XenopoulosS21}. We provide benchmarks for two game state representations. First, we provide benchmarks for vector representations of $g_t$. In this setup, we define $g_{t}$ as a collection of global and team-specific information, based on past literature~\cite{DBLP:conf/bigdataconf/XenopoulosDS20, DBLP:conf/www/XenopoulosFS22}, namely: the time since the last game phase change, total number of active fires and smokes, where the bomb is planted (A, B, or None), the total number of defuse kits on alive players, the starting equipment values of each side and each side's total number of alive players, current equipment value, starting equipment value, HP, armor, helmets, grenades remaining, and count of players in a bomb zone. Additionally, we also include a flag indicating if the bomb is in a T-side player's inventory. Notably, these vector representations lack player position information, since player-specific information would need to be presented in a permutation invariant manner, such as through an aggregation (e.g., mean or sum). While one could consider more complex representations of a game state, using a single vector with both global and aggregated team information is a common approach in many sports. Thus, it is an important game state representation to benchmark against.

We also provide benchmarks for models that consider $g_t$ as a set, where each $p_{i,t} \in g_t$ represents the information of player $i$ at time $t$. Each $p_{i,t}$ contains global features like the time since the last game phase change, total number of active fires and smokes, where the bomb is planted (A, B, or None), as well as features broadly derived from the player information (location, velocity, view direction, HP, equipment), and flags indicating player-states, such as if the player is alive, blinded, or in the bomb zone. A set representation of $g_t$ has yet to be used for win probability prediction in any sport. 

\subsection{Models and Training Setup}
Boosted tree ensembles are commonly used for win probability prediction due to their ease-of-use, strong performance, and built-in feature importance calculations. For vector representations of $g_t$, we consider XGBoost~\cite{DBLP:conf/kdd/ChenG16}, LightGBM~\cite{DBLP:conf/nips/KeMFWCMYL17}, and a multilayer perceptron (MLP) as candidate models. For set representations of $g_t$, we consider Deep Sets~\cite{DBLP:conf/nips/ZaheerKRPSS17} and Set Transformers~\cite{DBLP:conf/icml/LeeLKKCT19}. For each parsed demo, we randomly sample one game state from each round. Then, we randomly select 70\% of these game states for the train set, 10\% for the validation set, and the remaining 20\% for the test set. The label for each game state is determined by the outcome of the round in which the game state took place (i.e., $Y_i = 1$ if the CT side won, 0 otherwise). All benchmarks are available in a Google Colab notebook and require a high-RAM instance with a GPU.

We separate our prediction tasks by map due to the unique geometry imposed by each map. From prior work, we know that CSGO maps often carry unique characteristics which influence win probability~\cite{DBLP:conf/www/XenopoulosFS22}. Furthermore, CSGO maps are not played at equal rates -- some maps, such as Mirage, Inferno, or Dust 2, are selected and played more often than maps like Vertigo or Ancient. Thus, we do not consider demos that occurred on the map Train due to its small sample size (52 games). In total, we have seven maps $\times$ five candidate models for a total of 35 benchmark models.

For LightGBM and XGBoost, we use the default parameters provided by their respective packages, as well as 10 early stopping rounds. For the MLP, we use the default sklearn parameters and scale each feature to be between 0 and 1. Our Deep Sets model uses one fully connected layer in the encoder and one fully connected layer in the decoder. We use the mean as the final aggregation. Our set transformer model uses one induced set attention block with one attention head for the encoder. For its decoder, we use a pooling-by-multihead-attention block~\cite{DBLP:conf/icml/LeeLKKCT19}. When training our Deep Sets and Set Transformer models, we use a batch size of 32 and a hidden vector size of 128 for all layers. We maximize the log likelihood of both models using Adam~\cite{DBLP:journals/corr/KingmaB14} with a learning rate of 0.001, and we train over 100 epochs with 10 early stopping rounds.

\begin{table}[]
\centering
\caption{Benchmark results by map, measured by log loss (LL) and calibration error (ECE). Results are averaged from 10 runs and reported with their standard error. The best log loss and calibration error for each map are in \textbf{bold}.}
\begin{tabular}{@{}ccccccc@{}}
\toprule
\multicolumn{2}{c}{\multirow{2}{*}{\textbf{Map}}}    & \multicolumn{3}{c}{\textit{Vector-based}}                          & \multicolumn{2}{c}{\textit{Set-based}} \\ \cmidrule(l){3-7} 
\multicolumn{2}{c}{} &
  \textbf{LightGBM} &
  \textbf{XGBoost} &
  \textbf{MLP} &
  \textbf{\begin{tabular}[c]{@{}c@{}}Deep\\ Sets\end{tabular}} &
  \textbf{\begin{tabular}[c]{@{}c@{}}Set\\ Trans.\end{tabular}} \\ \midrule
\multirow{2}{*}{Dust2}    & \multicolumn{1}{c|}{LL}  & 0.433±0.005          & 0.437±0.005          & \textbf{0.415±0.003} & 0.435±0.005  & 0.458±0.007             \\
                          & \multicolumn{1}{c|}{ECE} & 0.042±0.002          & 0.044±0.002          & \textbf{0.034±0.003} & 0.047±0.004  & 0.043±0.003             \\ \midrule
\multirow{2}{*}{Mirage}   & \multicolumn{1}{c|}{LL}  & 0.439±0.004          & 0.440±0.004          & \textbf{0.430±0.004} & 0.440±0.003  & 0.447±0.003             \\
                          & \multicolumn{1}{c|}{ECE} & 0.041±0.001          & 0.039±0.002          & 0.038±0.004          & 0.040±0.002  & \textbf{0.035±0.004}    \\ \midrule
\multirow{2}{*}{Inferno}  & \multicolumn{1}{c|}{LL}  & 0.453±0.004          & 0.453±0.005          & \textbf{0.442±0.005} & 0.460±0.005  & 0.472±0.005             \\
                          & \multicolumn{1}{c|}{ECE} & 0.040±0.002          & \textbf{0.036±0.003} & 0.038±0.002          & 0.043±0.002  & 0.041±0.003             \\ \midrule
\multirow{2}{*}{Nuke}     & \multicolumn{1}{c|}{LL}  & 0.429±0.004          & 0.432±0.004          & \textbf{0.416±0.004} & 0.427±.004   & 0.447±0.006             \\
                          & \multicolumn{1}{c|}{ECE} & 0.037±0.002          & \textbf{0.033±0.003} & 0.038±0.001          & 0.036±0.001  & 0.038±0.003             \\ \midrule
\multirow{2}{*}{Overpass} & \multicolumn{1}{c|}{LL}  & 0.457±0.004          & 0.460±0.004          & \textbf{0.438±0.005} & 0.452±0.004  & 0.470±0.007             \\
                          & \multicolumn{1}{c|}{ECE} & \textbf{0.040±0.002} & 0.043±0.004          & 0.040±0.004          & 0.040±0.003  & 0.043±0.004             \\ \midrule
\multirow{2}{*}{Vertigo}  & \multicolumn{1}{c|}{LL}  & 0.442±0.006          & 0.441±0.005          & \textbf{0.424±0.005} & 0.441±0.005  & 0.456±0.005             \\
                          & \multicolumn{1}{c|}{ECE} & 0.048±0.002          & 0.045±0.002          & 0.046±0.002          & 0.046±0.002  & \textbf{0.040±0.003}  \\ \midrule
\multirow{2}{*}{Ancient}  & \multicolumn{1}{c|}{LL}  & 0.456±0.009          & 0.457±0.009          & \textbf{0.431±0.008} & 0.458±0.007  & 0.470±0.006             \\
                          & \multicolumn{1}{c|}{ECE} & 0.061±0.005          & 0.055±0.002          & 0.052±0.003          & 0.051±0.003  & \textbf{0.047±0.003}    \\ \bottomrule
\end{tabular}
\label{tab:benchmark-results}
\end{table}

\subsection{Assessing Win Probability Models}
To assess our benchmarked models, we use log loss (LL) and expected calibration error (ECE). ECE is a quantitative measurement of model calibration, and attempts to measure how closely a model's predicted probabilities track true outcomes. ECE is a common metric to assess win probability models since for many applications, stakeholders use the probabilities directly~\cite{DBLP:conf/bigdataconf/XenopoulosDS20, DBLP:conf/www/XenopoulosFS22}. Formally, ECE is defined as 

\begin{equation}
    ECE = \sum_{w=1}^W \frac{|B_w|}{N} | \textrm{acc}(B_w) - \textrm{conf}(B_w) |
\end{equation}

\noindent where $W$ is the number of equal width bins between 0 and 1, $B_w$ is the set of points in bin $w$, $\textrm{acc}(B_w)$ is the true proportion of the positive class in $B_w$, $\textrm{conf}(B_w)$ is the average prediction in $B_w$, and $N$ is the total number of samples. When assessing our candidate models, we set $W = 10$.

\subsection{Results}
We report the results of the log loss and expected calibration error by CSGO map and model in Table~\ref{tab:benchmark-results}. No model performed the best across all maps and metrics. We note that, in general, the models considering vector-based input, such as LightGBM, XGBoost, and in particular, the MLP achieved strong performance against our set-based approaches. Concerning the set-based models, we observed that the Deep Sets slightly outperformed the Set Transformer.

% One concern may be that, since we only sample one state from each round, that we may not have a large enough sample size for our Deep Sets and Set Transformer models. To test this, we instead split by round into train/validation/test sets, rather than by state. Then, we sampled every game state from each round. However, we found limited improvement, if any, to increasing sample size by considering all game states. 

\begin{figure}
    \centering
    \includegraphics[width=\linewidth]{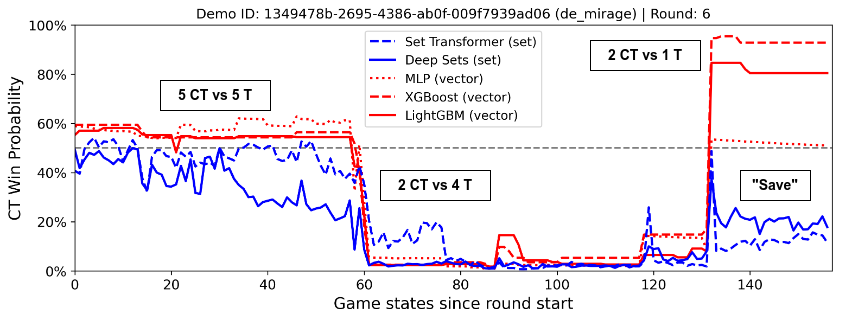}
    % save_example.png
    % round_save_example.png
    % Dust 2 map
    \caption{Win probability predictions for a CSGO round, colored by model. We find that in general, the models capture the effects of important game events, such as player damage events or the bomb plant. Although the set-based models perform slightly worse in terms of log loss, they are able to accurately account for player location-based game phenomena, such as the ``save'', whereby the CT side elects to avoid defusing a planted bomb.}
    \label{fig:win-prob-example}
\end{figure}

In Figure~\ref{fig:win-prob-example}, we show an example of each model's predicted win probability for a CSGO round on the Mirage map. We see that, although our game state formulation is simple, each of the models is able to capture the impact of important game events, such as damages. These events have clear impacts on the round, and especially so for the features with the highest importance in the gradient boosting models, such as each team's HP or if the bomb is planted. Otherwise, we also notice that the gradient boosted tree models have relatively ``flat'' predictions in certain intervals, which is sensible given that the gradient boosted models should return non-smooth decision surfaces.

During the round shown in Figure~\ref{fig:win-prob-example}, there is a large fight around the 60\textsuperscript{th} game state which causes the game situation to change from 5 CT versus 5 T to 2 CT versus 4 T. Although the bomb is planted around the 80\textsuperscript{th} frame, the effect is small given the already low prior predicted win probability. When the bomb is planted in CSGO, the CT side must decide whether to attempt to defuse the bomb or to ``save'', meaning the team elects to purposely lose the round by avoiding a bomb defusal (thus, forcing their ``true'' win probability to 0). The reason for the latter strategy is so that the saving team keeps their equipment for the next round, so they do not have to spend money. In the situation posed in Figure~\ref{fig:win-prob-example}, the bomb is planted when there are 2 CT players and 4 T players remaining. However, the CT players decide to save in this instance. In the process of saving, they eliminate 3 T players. We can see that the models diverge past the 130\textsuperscript{th} game state, and that, due to the perceived 2-on-1 scenario, the vector-based models produce high win probability predictions. These predictions run counter to conventional game knowledge. Conversely, the set-based models, which account for player locations, are less sensitive to such an increase, which aligns with conventional game knowledge. 

\section{Concluding Remarks} \label{sec:discussion}
% talk about improving representations of game states, considering sequences
% multi-agent blah blah
% Each JSON is highly granular and can be used to reconstruct the game as it happened.
We present ESTA, a large-scale and granular dataset containing millions of player actions and trajectories from professional CSGO matches. ESTA is made available via Github and contains roughly 1.5k compressed JSONs, where each JSON represents one game replay. We create ESTA using \texttt{awpy}, an open-source Python library which can parse, analyze, and visualize Counter-Strike game replays. Additionally, we provide benchmarks for win probability prediction, a foundational sports-specific prediction task. As part of these benchmarks, we apply set learning techniques, such as Set Transformers, to win probability prediction. Although the set-based techniques do not outperform the vector-based models, they are potentially able to elucidate some high-level game phenomena. We do not foresee any negative societal impact of our work.

\textbf{Limitations}. While we have made an effort to address limitations during the creation of ESTA, there may still be some constraints in using ESTA. First, while CSGO demos are often recorded at 64 or 128Hz, we record game states at 2Hz. While this may impose difficulties in understanding fine-grained movements, CSGO players generally do not move much within the space of one second. Additionally, the demos were recorded on servers with third-party plugins, which are common for servers used in tournaments. Generally, these plugins help facilitate the CSGO match. For example, in the rare event that a player disconnects during a round, the round will need to be restarted, and these server plugins can help reset the server to a prior state. While the awpy package does contain functionality to clean out such occurrences, some situations may not be fully cleaned by the parser. However, the number of rounds in the final parsed demo has been verified with the true number of rounds, so it is unlikely that such a problem is pervasive in ESTA.

% Our benchmarks suggest that conventional gradient-boosted tree methods to predict win probability work well and provide a competitive benchmark. Thus, future directions to predict win probability may focus on improved representations of the spatial characteristics of players, or incorporate past game state information, such as through sequence-based representations of game states~\cite{yurko2020going}. 

\textbf{Uses of ESTA}. ESTA has multiple uses beyond those benchmarked and portrayed in this work, not only in sports-specific machine learning, but also to the general machine learning community. For example, ESTA provides a large amount of trajectories which one can use for trajectory prediction. Furthermore, unlike traditional sports trajectory data, the trajectory data in ESTA occurs on various maps, each with different geometry, which can provide multiple unique prediction tasks. ESTA can also be used for tasks outside of sports outcome prediction or trajectory prediction. For example, the use of reinforcement learning in combination with video games is well-documented~\cite{jaderberg2019human}. The ESTA dataset may be useful in priming the model for online learning. ESTA may also be of use to the visualization community, which works extensively with sports applications due to the multivariate, temporal, and spatial aspects of sports data which necessitate visual analytics solutions~\cite{DBLP:journals/cga/BasoleS16, ggviz, DBLP:conf/iwec/HorstZD21, DBLP:journals/csur/GudmundssonH17}. Thus, the same sports data challenges that affect the machine learning community, which the ESTA dataset addresses, also affect the visualization community. Finally, the ESTA dataset may also prove useful in education, as it is representative of real-world data and can be used for a variety of tasks.

% since its JSON format is similar to what students may see in real world applications. ESTA contains rich spatiotemporal data to be used for projects, and the size of the data used ca be controlled by selecting specific tournaments. 

\bibliographystyle{plain}
\bibliography{00_references}

\medskip

%%%%%%%%%%%%%%%%%%%%%%%%%%%%%%%%%%%%%%%%%%%%%%%%%%%%%%%%%%%%

% \input{checklist}

%%%%%%%%%%%%%%%%%%%%%%%%%%%%%%%%%%%%%%%%%%%%%%%%%%%%%%%%%%%%

% \appendix
% \input{appendix}

\end{document}